\documentclass{article} 
\usepackage[dvipsnames]{xcolor}
\usepackage{inconsolata}
\usepackage[final]{colm2025_conference}
\usepackage{microtype}
\usepackage{hyperref}
\usepackage{url}
\usepackage{booktabs}
\usepackage[none]{hyphenat}
\usepackage{tabularx}
\usepackage{lineno}
\usepackage{tabularray}
\usepackage{tabu}
\usepackage{multirow}
\usepackage{graphicx}
\usepackage{svg}
\usepackage{adjustbox}
\usepackage{tikz}
\usepackage{listings}
\usepackage{xcolor}
\usepackage{fvextra}     
\usepackage{tcolorbox}   

\definecolor{darkblue}{rgb}{0, 0, 0.5}
\hypersetup{colorlinks=true, citecolor=darkblue, linkcolor=darkblue, urlcolor=darkblue}

\title{ModernBERT or DeBERTaV3? Examining Architecture and Data Influence on Transformer Encoder Models Performance}

\author{Wissam Antoun, Benoît Sagot \& Djamé Seddah \\
Inria, Paris, France\\
\texttt{\{wissam.antoun,benoit.sagot,djame.seddah\}@inria.fr} \\
}

\begin{document}

\ifcolmsubmission
\linenumbers
\fi

\maketitle

\begin{abstract}
Pretrained transformer-encoder models like DeBERTaV3 and ModernBERT introduce architectural advancements aimed at improving efficiency and performance. 
Although the authors of ModernBERT report improved performance over DeBERTaV3 on several benchmarks, the lack of disclosed training data and the absence of comparisons using a shared dataset make it difficult to determine whether these gains are due to architectural improvements or differences in training data.
In this work, we conduct a controlled study by pretraining ModernBERT on the same dataset as CamemBERTaV2, a DeBERTaV3 French model, isolating the effect of model design. 
Our results show that the previous model generation remains superior in sample efficiency and overall benchmark performance, with ModernBERT’s primary advantage being its support for long context, faster training, and inference speed.
However, the new proposed model still provides meaningful architectural improvements compared to earlier models such as BERT and RoBERTa.
Additionally, we observe that high-quality pre-training data accelerates convergence but does not significantly improve final performance, suggesting potential benchmark saturation. 
These findings show the importance of disentangling pretraining data from architectural innovations when evaluating transformer models.
\end{abstract}

\section{Introduction}
Despite the widespread adoption of decoder-only large language models (LLMs) in our post-ChatGPT era, encoder-only transformers such as BERT~\citep{devlin-etal-2019-bert} continue to play a central role in many NLP applications. 
These models remain the backbone of a wide range of non-generative tasks such as classification, named entity recognition (NER), and retrieval-based systems, especially in high-throughput or latency-sensitive environments. 
Their relatively low compute requirements and strong performance across standard information benchmarks benchmarks make them a practical choice for large-scale deployment, including in Retrieval-Augmented Generation~\citep{lewis2020retrieval} pipelines or guardrails systems~\citep{neill2024guardformer}.
Notable examples include Google’s EmbeddingGemma~\citep{vera2025embeddinggemma}, the BGE family of 33 models~\cite{bge-m3,bge_embedding}, and multilingual encoders such as multi-ligual modernBERT~\cite{marone2025mmbert} and EuroBERT~\cite{boizard2025eurobert}, alongside GTE-ModernBERT~\cite{li2023towards,zhang2024mgte}.

Continuous architectural and training objective improvements have led to more performant and efficient encoder-only transformer variants, among which DeBERTaV3~\citep{he2021debertav3} and the recently proposed ModernBERT~\citep{warner2024smarter} stand out as major improvements.
According to \citet{warner2024smarter}, their model reports superior performance relative to DeBERTaV3, the previous state-of-art model, across several popular NLP benchmarks. 
However, interpreting these performance improvements is challenging due to the lack of details regarding their training data. 
Without comparisons conducted on identical datasets, it remains unclear whether the reported gains reflect genuine architectural advances or simply differences arising from the choice of training data.

This uncertainty motivates our study aimed to evaluate the impact of architecture versus training data by conducting a controlled comparison among ModernBERT, DeBERTaV3, and RoBERTa models. 
We select CamemBERTaV2~\citep{antoun2024camembert,antoun2023data}, a French DeBERTaV3 model trained from scratch, as our primary reference since both its intermediate checkpoints and training dataset are publicly available. 
Additionally, we include CamemBERTv2~\citep{antoun2023data, antoun2024camembert}, a RoBERTa~\citep{liu2019roberta} based model pretrained on the same dataset, to comprehensively assess how ModernBERT's architectural advancements compare not only against the latest DeBERTaV3-based models but also against more traditional BERT/RoBERTa architectures. 
Leveraging these resources, we pretrained a French ModernBERT using the exact same dataset as CamemBERTaV2 and CamemBERTv2, thus ensuring that differences in performance directly reflect architectural variations rather than dataset composition or quality. 
In addition, we pretrained another ModernBERT variant on a carefully curated, high-quality French corpus to further explore the role of dataset quality in model performance.

The key takeaways from our comprehensive experiments are as follows:
\begin{itemize}

    \item When controlling for dataset differences, DeBERTaV3 outperforms ModernBERT in terms of overall benchmark performance and training sample efficiency, with the notable exception of text retrieval tasks where DeBERTaV3 fails completely. This indicates that while DeBERTaV3's architecture and training objective optimization provide superior learning capabilities compared to ModernBERT’s efficiency-oriented design, they do not generalize to all tasks.

    \item Nonetheless, ModernBERT presents a clear practical advantage due to its significantly faster training and inference speeds, driven by an orthogonal set of optimizations. Moreover, while ModernBERT may not surpass DeBERTaV3, it offers meaningful improvements over previous transformer-based models such as BERT and RoBERTa.

    \item We also observe that training models on high-quality, filtered datasets results in faster convergence but does not substantially increase final performance metrics. This finding highlights a potential limitation of current NLP benchmarks, suggesting possible saturation that prevents fine-grained discrimination between models of similar performance.
    
\end{itemize}

Our findings show the importance of clearly separating respective effects of architectural changes and training datasets when evaluating NLP models.
Our controlled comparison using the same pretraining dataset provides more accurate insights into the strengths and limitations of ModernBERT and DeBERTaV3 architectures.

To promote further research and ensure reproducibility, we publicly release our two pretrained French ModernBERT models, collectively named \textbf{ModernCamemBERT}, including one trained on the CamemBERTaV2 dataset and another on our high-quality filtered corpus. These models, along with intermediate checkpoints, and evaluation results, are available on HuggingFace\footnote{\href{https://huggingface.co/collections/almanach/moderncamembert}{https://huggingface.co/collections/almanach/moderncamembert}}.

\section{Related Works}
Transformer-based language models have become the cornerstone of modern NLP, starting with BERT~\citep{devlin-etal-2019-bert}, which introduced masked language modeling (MLM) and next sentence prediction as self-supervised pretraining tasks used to pretrain encoder-only transformer models.
RoBERTa~\citep{liu2019roberta} subsequently improved upon BERT by removing the next sentence prediction objective, training on larger corpora, and applying more robust optimization techniques.

Despite these advances, both BERT and RoBERTa shared a fundamental architectural limitation: they used absolute positional embeddings and standard attention mechanisms that lacked efficiency and fine-grained contextual representation.
To address these limitations, DeBERTa~\citep{he2021deberta} introduced a disentangled attention mechanism, decoupling content and positional information, thereby improving the model’s ability to generalize across contexts. 
DeBERTaV3~\citep{he2021debertav3} further extended these innovations by incorporating Replaced Token Detection (RTD)~\citep{clark2020electra} for more sample-efficient training, as well as Gradient-Disentangled Embedding Sharing (GDES) to prevent conflicting updates in shared embeddings between the generator and discriminator during training.

In parallel, architectural and efficiency-driven improvements have become an active area of research. ModernBERT~\citep{warner2024smarter} was recently proposed to modernize the BERT architecture by incorporating a suite of design choices aimed at improving inference speed, training throughput, and context window size. 
These include FlashAttention~\citep{dao2022flashattention, dao2024flashattention, shah2024flashattention}, alternating global and local attention layers~\citep{team2024gemma}, sequence packing~\citep{portes2023mosaicbert}, and rotary positional embeddings (RoPE)~\citep{su2021roformer}. 
ModernBERT also removes architectural elements such as bias terms and introduces GeGLU~\citep{shazeer2020glu} activations, making it a strong contender for production scenarios requiring high efficiency.

While ModernBERT has gained popularity other models have also made significant contributions to the field.
For instance, MosaicBERT~\cite{portes2023mosaicbert} was the first to focus on enhancing training efficiency and performance by increasing masking rates, sequence packing and FlashAttention.
NomicBERT~\cite{nussbaum2025nomicembedtrainingreproducible} introduced architectural improvements like SwiGLU activation functions, RoPE positional encoding and extended context lengths, enhancing its ability to handle longer sequences up to 2048 tokens.
NeoBERT~\cite{breton2025neobertnextgenerationbert} further advanced these developments by optimizing the depth-to-width ratio and doubling the context length, while also significantly increasing training corpus size.

We chose to compare against ModernBERT because it was the best available model at the moment we started our experimentation and it is the most popular encoder in the field.
While the authors of ModernBERT report superior benchmark performance over DeBERTaV3, the absence of transparent training data and lack of head-to-head comparisons on shared datasets introduces ambiguity. It is thus unclear whether the reported improvements are driven by architectural enhancements or the underlying training data.
\section{Methodology}

We conduct a controlled study focusing on the performance of ModernBERT compared to DeBERTaV3-based and RoBERTa-based models.
Our goal is to identify and separate architectural improvements from data-driven performance differences, addressing ambiguities in prior studies that used undisclosed datasets.

\subsection{Pre-training Datasets}

Our experiments involve two distinct pre-training datasets:

\paragraph{CamemBERTaV2 Original Dataset.}
We first make use of the publicly available French dataset originally used by the authors of CamemBERTaV2~\citep{antoun2024camembert}.
This dataset has 275 billion tokens, sourced from:

\begin{itemize}
    \item \textbf{CulturaX-FR Corpus}~\citep{nguyen2023culturax}: The French subset of a multilingual corpus containing around 265 billion French tokens, constructed from multiple snapshots of OSCAR~\cite{OrtizSuarezSagotRomary2019,AbadjiOrtizSuarezRomaryetal.2021,abadji-etal-2022-towards}  and mC4~\cite{xue-etal-2021-mt5} corpora.
    \item \textbf{HALvesting Corpus}~\citep{kulumba2024harvestingtextualstructureddata}: Approximately 4.7 billion tokens of academic and scientific content from French theses and research papers.
    \item \textbf{French Wikipedia}: Roughly 0.5 billion tokens from Wikipedia, intentionally upsampled to enhance general knowledge representation.
\end{itemize}

This dataset serves as a reference point, allowing us to directly compare models under identical training dataset conditions.

\paragraph{High-Quality Filtered Dataset.}
We also feature a second significantly larger 1T tokens French dataset created by applying heuristic and semantic filters in addition to full deduplication to the French section of the RedPajamaV2 corpus~\citep{weber2024redpajama}, combined with the HALvesting corpus and French Wikipedia.
Semantic filtering was done following the FineWeb-Edu~\citep{penedo2024finewebdatasetsdecantingweb} methodology.
This method has been effective in increasing the overall quality of a corpus used for LLM training and has been widely adopted in the literature~\cite{li2024datacomp,su2025nemotron}.
First, We annotated 200K samples from the RedPajamaV2 dataset with quality labels (low, medium, high), using the LLama-3 70B model~\citep{llama3} and the prompt provided in Appendix~\ref{appendix:prompt}.
This annotated subset was then used to fine-tune XLM-V-base~\citep{liang2023xlmvovercomingvocabularybottleneck}, which we use to annotate the whole RedPajamaV2 corpus.
The semantic score was combined with the perplexity score from a language model trained on Wikipedia, as included in the RedPajama dataset. The RedPajama authors categorize data into head, middle, and tail buckets based on perplexity.
We only select data if it is either from the head bucket or has a high score from the quality classifier, while disgrading any tail or low labeled documents.

\subsection{Model Training}

We pretrained two variants of the ModernBERT model, one on the CamemBERTaV2 dataset, which we call ModernBERT-CV2, and the other on our high-quality filtered dataset, named ModernBERT-HQ, periodically saving intermediate checkpoints.\footnote{We use the publicly available ModernBERT codebase~\href{https://github.com/AnswerDotAI/ModernBERT}{https://github.com/AnswerDotAI/ModernBERT}}
Both variants are base-sized models trained with Masked Language Modeling (MLM)  for 1 trillion tokens, with a maximum sequence length of 1024, and use the same tokenizer as CamemBERTaV2.
We maintained the rate of dynamic token masking to 30\%, while retaining all other hyperparameters consistent with those of ModernBERT-base.
Training was done on 48 NVidia H100 80GB GPUs using Pytorch's FSDP full sharding with bfloat16 mixed precision to speed up training\footnote{See pretraining hyperparameter details in Appendix~\ref{appendix:pretraining-hp}}.

Since our models were trained using a Warmup-Stable-Decay (WSD) learning rate schedule, each intermediate checkpoint underwent additional cooldown training over an extra 50 billion tokens, during which the learning rate decayed fully to zero, ensuring fair comparisons across checkpoints.

Additionally, we leveraged publicly available intermediate checkpoints from the CamemBERTaV2 and CamemBERTv2 models, allowing direct comparisons of learning trajectories and data efficiency across different architectures. 
\begin{table*}[!t]
\centering
\resizebox{\textwidth}{!}{
    \begin{tabu}{lccccccc}
        \toprule
        & \textsc{NER} & \multicolumn{2}{c}{\textsc{QA}} & \textsc{CLS} & \textsc{PAWS-X} & \textsc{XNLI} & \textsc{MTEB} \\
        \cmidrule(lr){2-2}\cmidrule(lr){3-4}\cmidrule(lr){5-5}\cmidrule(lr){6-6}\cmidrule(lr){7-7}\cmidrule(lr){8-8}
        \multirow{-2.5}{*}{\textsc{Model}} & \textsc{F1}                             & \textsc{F1}                             & \textsc{EM}                             & \textsc{Acc}                            & \textsc{Acc}                            & \textsc{Acc}                            & \textsc{Avg}                            \\
        \midrule
        CamemBERTV2                        & 91.99{\scriptsize$\pm$0.96}             & 80.39{\scriptsize$\pm$0.36}             & 61.35{\scriptsize$\pm$0.39}             & 95.07{\scriptsize$\pm$0.11}             & 92.00{\scriptsize$\pm$0.24}             & 81.75{\scriptsize$\pm$0.62}             & \textbf{51.67{\scriptsize$\pm$0.57}}\\
            \midrule
        CamemBERTaV2                       & \textbf{93.40{\scriptsize$\pm$0.62}}    & \textbf{83.04{\scriptsize$\pm$0.19}}    & \textbf{64.29{\scriptsize$\pm$0.31}}    & \textbf{95.63{\scriptsize$\pm$0.16}}    & \textbf{93.06{\scriptsize$\pm$0.45}}    & \textbf{84.82{\scriptsize$\pm$0.54}}    & 31.15{\scriptsize$\pm$5.26}\\
        \midrule
        ModernBERT-CV2                     & \underline{92.03{\scriptsize$\pm$0.14}} & \underline{81.34{\scriptsize$\pm$0.35}} & 61.47{\scriptsize$\pm$0.46}             & \underline{95.18{\scriptsize$\pm$0.20}} & \underline{92.79{\scriptsize$\pm$0.22}} & 83.28{\scriptsize$\pm$0.34}             & 49.44{\scriptsize$\pm$1.36}\\
        ModernBERT-HQ                      & 91.80{\scriptsize$\pm$0.47}             & 81.11{\scriptsize$\pm$0.26}             & \underline{62.07{\scriptsize$\pm$0.44}} & 95.04{\scriptsize$\pm$0.09}             & 92.55{\scriptsize$\pm$0.54}             & \underline{83.66{\scriptsize$\pm$0.67}} & \underline{49.93{\scriptsize$\pm$0.60}}\\
        \bottomrule
    \end{tabu}
}
\caption{Downstream tasks results. \textbf{Bold} indicates best score overall while \underline{underline} indicates best score between the ModernBERT models. \textbf{ModernBERT-CV2} is the ModernBERT model trained on the same data as CamemBERTaV2 while \textbf{ModernBERT-HQ} is the one trained on the high-quality filtered dataset. \textit{Scores are the 5-seed average of the best performing set of hyperparamters for each model. MTEB scores are the average over all tasks. Full MTEB scores are available in Table~\ref{tab:mteb_task_type_results}}}
\label{tab:downstream_all}
\end{table*}

\section{Experiments and Results}
\subsection{Downstream Evaluation Tasks}

To evaluate our models, we consider a range of French downstream tasks and datasets, including:
\begin{itemize}
    \item \textbf{Question Answering (QA)}: using FQuAD 1.0~\cite{2020arXiv200206071}
    \item \textbf{Named Entity Recognition (NER)}: on the 2008 FTB version~\citep{abeille-etal-2000-building, candito-crabbe-2009-improving} with NER annotations by Sagot et al.~\citep{sagot-etal-2012-annotation}
    \item \textbf{Text Classification} capabilities assessed using the FLUE benchmark~\citep{le-etal-2020-flaubert-unsupervised} using the CLS amazon reviews classification task, the PAWS-X paraphrase identification task and XNLI task.
    \item \textbf{Text Retrieval}: We used the French subset of the translated Semantic Textual Similarity (STS) benchmark~\citep{huggingface:dataset:stsb_multi_mt} for training and then evaluated the resulting models using the French Massive Text Embedding Benchmark (MTEB)~\citep{ciancone2024mtebfrench,enevoldsen2025mmtebmassivemultilingualtext,muennighoff2022mteb}.

\end{itemize}

We re-used the same splits from the CamemBERTaV2 authors and performed hyper-parameter tuning on all models and datasets with 5 seed variations.

\subsection{Downstream Results Analysis}

The downstream evaluation results summarized in Table~\ref{tab:downstream_all} show the following insights into model architectures and pretaining dataset effects:

\paragraph{Architecture Impact.} Comparing the models trained on identical datasets (ModernBERT-CV2 and CamemBERTaV2/CamemBERTv2), we observe that ModernBERT-CV2 consistently outperforms CamemBERTv2 with the exception of text retrieval, thus showing ModernBERT's improvements over BERT/RoBERTa.
However, it fails to surpass CamemBERTaV2 on any non-retrieval task, even though the latter being only trained for a single epoch on the dataset compared to three epochs (1T tokens) for ModernBERT-CV2.
This clearly demonstrates that while ModernBERT offers valuable throughput-driven architectural enhancements, these improvements do not match the contextual learning capabilities provided by DeBERTaV3's disentangled attention and RTD-based pretraining objective.
Our results also confirm the observation that DeBERTaV3 fails on text embedding tasks.
Despite its strong performance on NLU tasks, its sentence representations are poorly structured in the embedding space.

\paragraph{Data Quality Impact.} Interestingly, switching to our high-quality filtered dataset (ModernBERT-HQ) only marginally improved performance on downstream tasks, despite the dataset containing three times more unique tokens than the original CamemBERTaV2 dataset.
ModernBERT-HQ slightly outperformed ModernBERT-CV2 on QA (FQuad), CLS, XNLI and text retrieval tasks, but improvements remained within small margins. 
This limited gain suggests two potential explanations: either current transformer architectures exhibit diminishing returns when exposed to additional data beyond a certain threshold, or standard French benchmarks are becoming saturated and unfit to measure model quality with further improvements in model performance. 
The latter possibility stresses the need for more challenging and diverse benchmarks that can effectively capture the improvements brought by higher-quality data.

\subsection{Pre-training Dynamics and Sample Efficiency}
We further explored the learning trajectories of the various models by evaluating intermediate checkpoints on QA (FQuad) and NER tasks. 
This analysis offers a more detailed view of the training dynamics and sample efficiency:

\begin{figure}[t]
    \centering
    \includegraphics[width=0.8\columnwidth, trim={0.9cm 0cm 2cm 2.5cm}, clip]{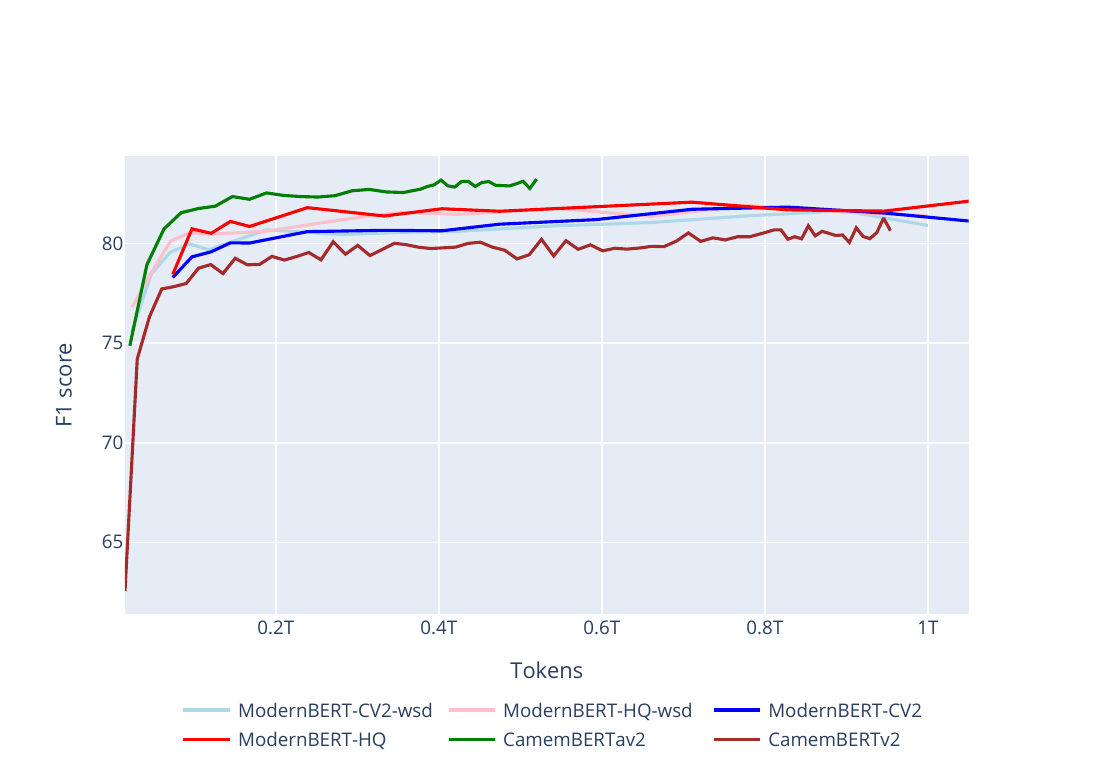}
    \caption{Downstream Performance on QA throughout the pre-training stage. \textit{wsd} are the models tested before the cooldown period.}
    \label{fig:evolution_qa}
\end{figure}

\begin{figure}[t]
    \centering
    \includegraphics[width=0.8\columnwidth, trim={0.9cm 0cm 2cm 2.5cm}, clip]{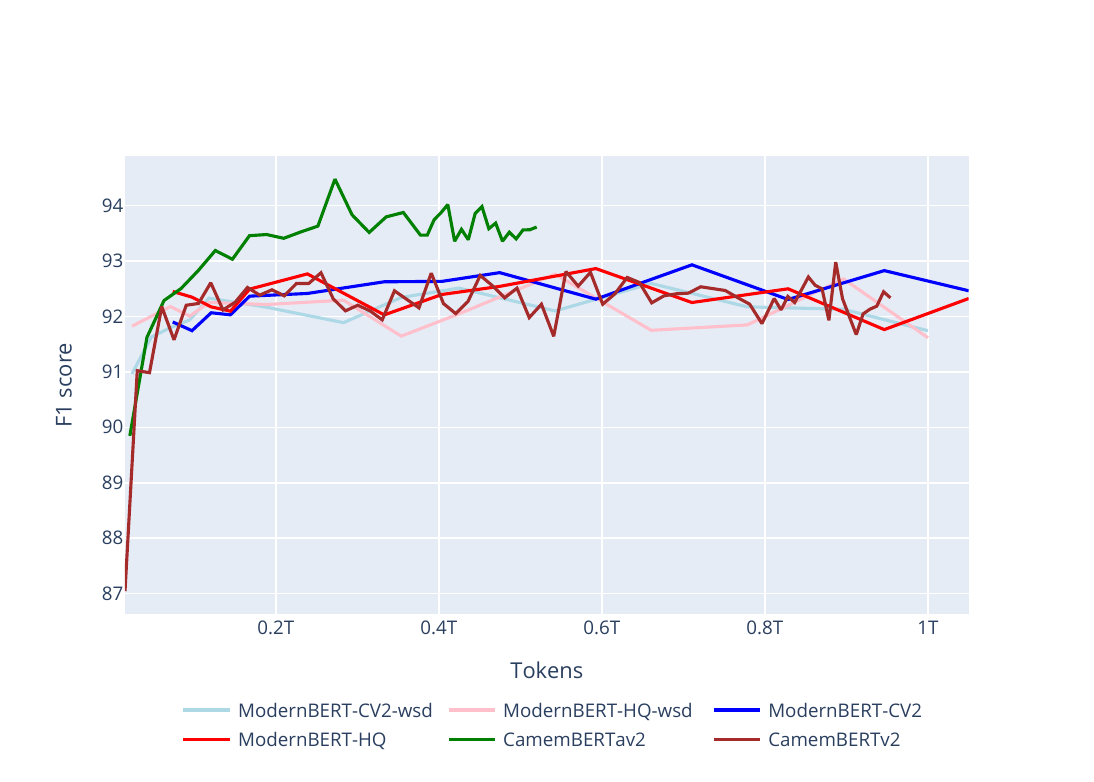}
    \caption{Downstream Performance on NER throughout the pre-training stage. \textit{wsd} are the models tested before the cooldown period.}
    \label{fig:evolution_ner}
\end{figure}

\DeclareRobustCommand{\thickgreenuparrow}{\raisebox{-1.5pt}{\hspace{4pt}\tikz \draw[->, ForestGreen, line width=1.2pt] (0,0) -- (0,0.3);\hspace{4pt}}}
\DeclareRobustCommand{\thickreddownarrow}{\raisebox{-1.5pt}{\hspace{4pt}\tikz \draw[->, BrickRed, line width=1.2pt] (0,0.3) -- (0,0);\hspace{4pt}}}
\DeclareRobustCommand{\yellowdash}{\raisebox{2pt}{\hspace{1pt}\tikz \draw[-, Goldenrod, line width=1.5pt] (0,0) -- (0.2,0);\hspace{1pt}}}

\begin{table}[t]
	\centering
	\resizebox{\textwidth}{!}{
		\begin{tabu}{lccccccc}
			\toprule
			& \textsc{NER} & \multicolumn{2}{c}{\textsc{QA}} & \textsc{CLS} & \textsc{PAWS-X} & \textsc{XNLI} & \textsc{MTEB} \\
			\cmidrule(lr){2-2}\cmidrule(lr){3-4}\cmidrule(lr){5-5}\cmidrule(lr){6-6}\cmidrule(lr){7-7}\cmidrule(lr){8-8}
			\multirow{-2.5}{*}{\textsc{Model}} & \textsc{F1}                                   & \textsc{F1}                                   & \textsc{EM}                                   & \textsc{Acc}                                  & \textsc{Acc}                           & \textsc{Acc}                                  & \textsc{Avg}                                  \\
			\midrule
			ModernBERT-CV2                     & 92.03{\scriptsize$\pm$0.14}                   & 81.34{\scriptsize$\pm$0.35}                   & 61.47{\scriptsize$\pm$0.46}                   & 95.18{\scriptsize$\pm$0.20}                   & 92.79{\scriptsize$\pm$0.22}            & 83.28{\scriptsize$\pm$0.34}                   & 49.49{\scriptsize$\pm$1.36}                   \\
			ModernBERT-HQ                      & 91.80{\scriptsize$\pm$0.47}                   & 81.11{\scriptsize$\pm$0.26}                   & 62.07{\scriptsize$\pm$0.44}                   & 95.04{\scriptsize$\pm$0.09}                   & 92.55{\scriptsize$\pm$0.54}            & 83.66{\scriptsize$\pm$0.67}                   & 49.93{\scriptsize$\pm$0.60}                   \\
			\midrule
			ModernBERT-CV2-final               & 92.17{\scriptsize$\pm$0.48}\thickgreenuparrow & 81.68{\scriptsize$\pm$0.46}\thickgreenuparrow & 62.00{\scriptsize$\pm$0.53}\thickgreenuparrow & 94.86{\scriptsize$\pm$0.16}\thickreddownarrow & 92.71{\scriptsize$\pm$0.39}\yellowdash & 82.85{\scriptsize$\pm$0.45}\thickreddownarrow & 48.79{\scriptsize$\pm$0.45}\thickreddownarrow \\
			ModernBERT-HQ-final                & 91.33{\scriptsize$\pm$0.27}\thickreddownarrow & 82.19{\scriptsize$\pm$0.46}\thickgreenuparrow & 62.66{\scriptsize$\pm$0.79}\thickgreenuparrow & 94.92{\scriptsize$\pm$0.06}\thickgreenuparrow & 92.52{\scriptsize$\pm$0.36}\yellowdash & 83.62{\scriptsize$\pm$0.67}\yellowdash        & 49.29{\scriptsize$\pm$0.81}\thickreddownarrow \\
			\bottomrule
		\end{tabu}
	}
	\caption{Downstream tasks results after context extension and cooldown. \thickgreenuparrow, \thickreddownarrow , and \yellowdash  indicate an increase, decrease or no change in scores after continual pretraining. \textit{Scores are the 5-seed average of the best performing set of hyperparamters for each model.}}
	\label{tab:downstream_long_context}
\end{table}

\begin{table}[]
\centering
\resizebox{0.6\columnwidth}{!}{
\begin{tabular}{lccc}
\toprule
                          &  & \multicolumn{2}{c}{MLDR (NDCG@10)}           \\
                          \cmidrule(lr){3-4} 
\multirow{-2.5}{*}{\textsc{Model}} & \multirow{-2.5}{*}{\textsc{Context}} & Max & Avg   \\
\midrule
CamemBERTV2          & 1024        & 32.59       & 28.37{\scriptsize$\pm$2.77} \\
\midrule
CamemBERTaV2         & 1024        & 2.44        & 00.91{\scriptsize$\pm$1.08} \\
\midrule
ModernBERT-CV2       & 1024        & 21.45       & 10.39{\scriptsize$\pm$4.83} \\
ModernBERT-CV2-final & 8192        & 26.93       & 22.59{\scriptsize$\pm$2.73} \\
\midrule
ModernBERT-HQ        & 1024         & 31.76       & 25.80{\scriptsize$\pm$1.99} \\
ModernBERT-HQ-final  & 8192         & \textbf{39.07}       & \textbf{34.32}{\scriptsize$\pm$5.44} \\ 
\bottomrule
\end{tabular}
}
\caption{Maximum and highest 5-seed averaged NDCG@10 score for the Multi Long Doc Retrieval task.}
\label{tab:mldr}
\end{table}

\paragraph{Architectural Efficiency.} The training curves (shown in Figures~\ref{fig:evolution_qa} and ~\ref{fig:evolution_ner}) indicate that CamemBERTaV2 reaches higher performance significantly earlier in training compared to ModernBERT-CV2. 
The DeBERTaV3-based model's faster improvement rate strongly suggests its better sample efficiency is due to optimizations like RTD and gradient-disentangled embedding sharing (GDES). 
Moreover, in scenarios where pre-training data is limited or scarce, its architectures might be more advantageous.

\paragraph{Impact of Data Quality on Convergence.} When comparing ModernBERT-CV2 and ModernBERT-HQ downstream performance throughout the training on the challenging QA tasks~\ref{fig:evolution_qa}, we observed that the model trained on the higher-quality dataset achieved its performance plateau faster, indicating that improved data quality enhances training efficiency and accelerates convergence. 
Yet, it does not substantially increase the final task-specific performance scores, further confirming the hypothesis of saturation effects on standard NLP benchmarks.

\paragraph{Task-specific Dynamics.} The intermediate checkpoints downstream score shows a clear difference in learning dynamics between the QA and NER tasks. 
While QA scores continued to improve gradually throughout training for all models, NER performance plateaued relatively early, with minimal further improvements, except for the CamemBERTaV2 NER scores which increased steadily. 
This difference suggests that the disentangled attention mechanism, which separately encodes content and positional embeddings, provides an advantage on token-level tasks such as NER.

\subsection{Context Length Extension and Final Model Release}
One of ModernBERT's advantages is supporting longer context length due to it's more efficient attention implementation and alternation of local and global attention layers.
On the other hand, the older models had limited context length due to the high memory usage of their attention layer implementation.
Hence, in order to study the effect of context length extension, we  continue our ModernBERT's pretraining, as in the original model's strategy, and increase its maximum input length to 8,192 tokens. 
This phase also includes a cooldown stage, during which the learning rate is gradually reduced to zero over high-quality, long-context data.

To support this phase, we curated two dataset variants:

\begin{itemize} 
    \item \textbf{Long-Context Subset:} We filtered documents longer than 2,048 tokens and retained them fully. Shorter documents were retained with a 10\% probability to preserve some distributional diversity.

    \item \textbf{High-Quality Long-Context Subset:} For this version, we upsampled high-quality, long-form sources such as French Wikipedia and academic literature, while only retaining documents rated as "high quality" by our semantic filter within the HQ dataset. 
\end{itemize}

We resumed training for both model variants, the one trained on the CamemBERTaV2 dataset and our High-Quality dataset, using their corresponding long-context subsets. 
Training was done for an additional 150 billion tokens to extend context capabilities, using a fixed learning rate of $3\times10^{-4}$. 
This was followed by a final 100 billion token cooldown phase, during which the learning rate was linearly decayed to zero.

\paragraph{Impact on Downstream Performance.}
We observe in Table~\ref{tab:downstream_long_context}, that ModernBERT-CV2 provides modest gains in NER (+0.14 F1) and QA (+0.34 F1 / +0.53 EM), while performance slightly decreases on classification (CLS: –0.32 Acc, XNLI: –0.43 Acc), with PAWS-X remaining stable. 
Meanwhile, ModernBERT-HQ-final displays clear improvements in QA (+1.08 F1 / +0.59 EM) and CLS (+0.88 Acc), while maintaining stable results on PAWS-X and XNLI. 
Although NER and text retrieval performance drops slightly (–0.47 F1 and -0.64), the overall trend indicates that high-quality long-context pretraining primarily benefits tasks requiring deeper semantic understanding or longer-range dependencies.

To better understand the models' long context extrapolation, we evaluate all models trained on the French STS dataset from earlier on the French test subset of the Multi Long Doc Retrieval (MLDR) and present the results in Table~\ref{tab:mldr}. 
The results clearly demonstrate the importance of the long-context pretraining stage.
As expected from the MTEB evaluation, CamemBERTaV2 performs poorly on this retrieval benchmark, further highlighting the unsuitability of its architecture for sentence embedding tasks.
Both ModernBERT variants show remarkable improvements after the final training phase, with context-extended ModernBERT-HQ achieving the highest max and average score.

However, the performance of ModernBERT-CV2 (the ModernBERT variant trained on the original CamemBERTaV2 dataset) is unexpectedly poor, scoring significantly lower than both CamemBERTV2 and its counterpart trained on our high-quality dataset, ModernBERT-HQ.
We currently lack a definitive explanation for this gap.
One hypothesis relates to the fundamental differences between the two datasets used in our study.
The original CamemBERTaV2 dataset (275B tokens) consists primarily of web-crawled content from CulturaX-FR (constructed from OSCAR and mC4 snapshots).
In contrast, our high-quality filtered dataset (1T tokens) underwent extensive semantic filtering using FineWeb-Edu methodology, perplexity-based selection, and full deduplication on the RedPajamaV2 corpus, resulting in a more coherent and diverse collection of texts.
The original dataset's heavy reliance on statistical filters may lack the semantic coherence necessary for learning robust long-form representations, whereas our filtered dataset's emphasis on high-quality, diverse sources appears better suited for supporting complex semantic understanding required in retrieval tasks.

\subsection{Downstream Training Stability}
\begin{figure*}[tbh] 
    \centering 
    \includegraphics[width=0.9\textwidth, trim={0.9cm 0.5cm 2cm 2.5cm}, clip]{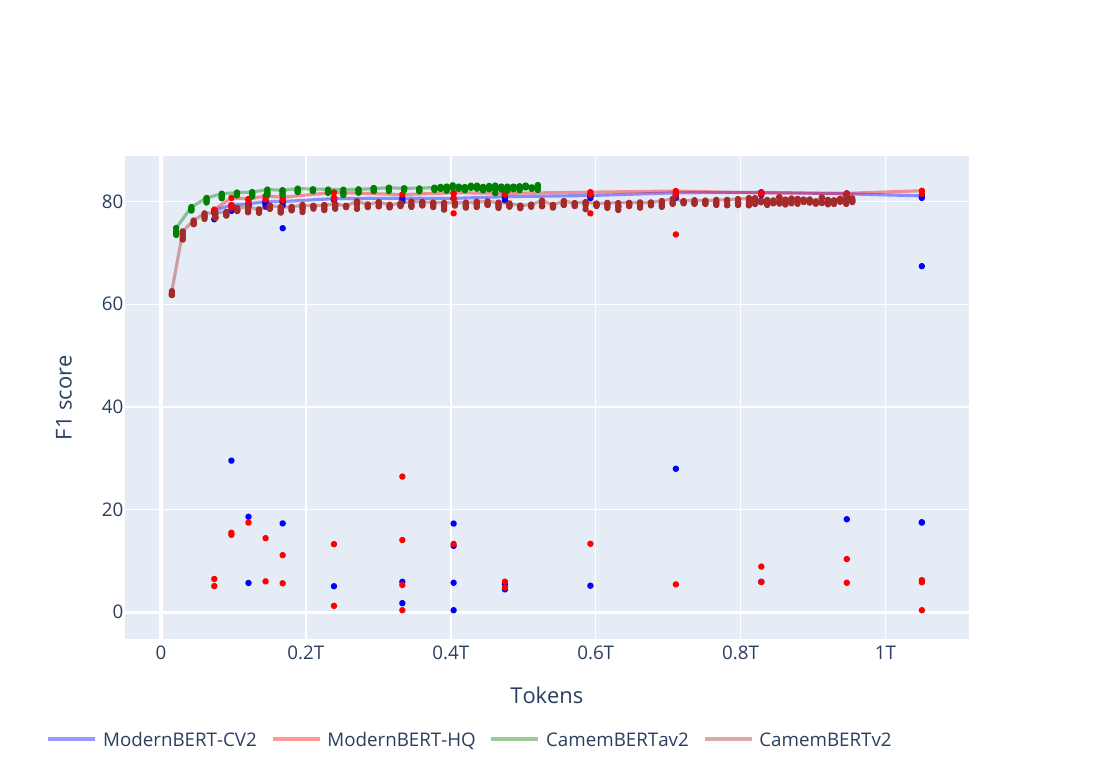} 
    \caption{Instances of divergence during QA fine-tuning. \textit{Colored lines illustrate the maximum score at a given step.}} \label{fig:instability_plot} 
\end{figure*}

During fine-tuning on downstream tasks, we observed differences in training stability between the newer and older model families. 
We had several cases where only ModernBERT variants failed to converge on the FQuAD question-answering task, as illustrated in Figure~\ref{fig:instability_plot}.

Furthermore, during hyperparameter tuning of the final checkpoint, we found the newer architecture to be particularly sensitive to learning rate choices.~\footnote{The search space included multiple learning rates (5e-5 to 5e-4), schedulers, batch sizes, and seeds.}
Despite hyperparameter tuning, the instability persisted. Further investigation revealed that NaN values appear in the loss from the very first training batch. This strongly supports our hypothesis of a numerical instability in the underlying implementation (e.g., FlashAttention) rather than a simple hyperparameter mismatch.


\subsection{Training Efficiency}
In addition to model accuracy, training efficiency is a crucial factor in practice, impacting both resource costs and environmental footprint.
For pretraining time, our ModernBERT training required 1300 H100 GPU-hours to complete one trillion tokens. 
On the other hand, CamemBERTv2 took roughly 2100 GPU-hours to train on the same dataset size while CamemBERTaV2 required around 2700 GPU-hours to complete just a single epoch, despite processing one-third of the tokens (275B). 
This clearly demonstrates ModernBERT's efficient architecture advantages and its practical edge during training.
However, it should be noted a significant portion of the speedup over DeBERTaV3-based models comes from engineering optimizations such as unpadding and FlashAttention, both of which are not implemented in the DeBERTa models at the time of this study.

The key takeaway from these experiments is the trade-off between ModernBERT, which offers significantly faster training and inference speeds, making it more efficient for time-sensitive applications, and DeBERTa, which delivers higher raw performance through its most effective use of training data.
\section{Discussion}

Confirming recent results~\cite{warner2024smarter,breton2025neobertnextgenerationbert}, our experiments showed a weakness in DeBERTaV3, which failed at the text retrieval task, despite its strong performance across other benchmarks. 
We hypothesize that its architecture may lack the mechanisms to build a global document representation suitable for retrieval using pooling of token embeddings. In contrast, ModernBERT performed well on this task, suggesting its alternating global and local attention layers provide a suitable architecture for text retrieval.

However, the strengths of ModernBERT in text retrieval did not extend to all tasks. Our experiments showed a deficiency in the ModernBERT question-answering experiments. We suspect that ModernBERT’s architecture, may face challenges in learning the long-range dependencies that are needed for QA tasks, particularly when informative signals are sparse.
For instance, the model needs to propagate the signal from the correct answer start/end token across long distances to the question context, which resides at the beginning of the input.
We hypothesize that while local attention layers are efficient, they might occasionally interrupt the direct propagation path for such sparse signals.
The global attention layers are then tasked with bridging these segments, leading to potential difficulties in predicting the answer span based on the distant question.
Other tasks such as NER, which rely on token-level prediction, do not require long-range dependencies since the model can infer the correct label based on local context, in addition to the richer learning signal, since named entity tokens are more frequent than a single start/end token per example.




Our work contributes to the renewed debate on encoder architectures versus the prevailing trend of using decoder-only models for all tasks (see \cite{Gisserot-Boukhlef_et_al:2005:Should_we_still} and references therein).
This discussion centers on whether a single, large architecture can be adapted for any task, or if specialized models remain superior for certain domains like NLU. 
The recent paper by~\citet{weller2025seqvsseqopen} offers new evidence by training paired encoder and decoder models under the same conditions. 
Their experiments confirm that encoders are better suited for classification and retrieval, which aligns with our results showing ModernBERT's strength in text retrieval. 
Furthermore, the paper demonstrates that adapting a model to a task for which it was not designed is an ineffective strategy. 
This puts our own results into a broader context, suggesting that specialized encoders are not obsolete, and that architectural choice remains a key factor for task performance.

Looking ahead, future work could explore integrating recent efficiency improvements such as \textit{FlashDeBERTa}\footnote{\href{https://github.com/Knowledgator/FlashDeBERTa}{https://github.com/Knowledgator/FlashDeBERTa}}, which applies FlashAttention to the disentangled attention mechanism and greatly reduces DeBERTa’s memory and latency costs. 
Another promising direction is investigating why DeBERTaV3 fails so strongly on embedding-based tasks. 
Examining its pooling mechanisms and representation geometry may help clarify the limits of disentangled attention for retrieval-oriented objectives.

\section{Conclusion}
We set out in this work to critically evaluate the claims made in the original ModernBERT paper by reproducing its setup under tightly controlled conditions. 
We isolated the authors' contributions by retraining their model under the same conditions as the previous state-of-the-art models, to assess their actual impact on training dynamics, efficiency, and downstream performance.

Our findings show that while ModernBERT does offer improvements in training and inference speed compared to older architectures, these do not translate into better sample efficiency or task performance under matched conditions. 
In fact, under careful evaluation, we found that DeBERTaV3's architecture and training objectives are more advantageous in low-data scenarios or if the goal is to get the absolute best task performance, except for the case of text retrieval where DeBERTaV3 fails completely.

We also observed that increasing the size and quality of pretraining data only yielded marginal gains for the newly proposed architecture, suggesting that current benchmarks may be reaching saturation, or at least they are insufficiently sensitive to capture finer improvements. 
During fine-tuning, we faced a problem with sensitivity to hyperparameters, which the V2 baselines did not have.
These stability concerns present challenges for reproducibility and deployment, and deserve further investigation.

In summary, ModernBERT offers a fast and efficient alternative for scenarios where training and inference speed are critical, but DeBERTaV3 remains the stronger choice when performance and sample efficiency are required.
Our study reinforces the importance of evaluating models under shared conditions to truly understand the contributions of architecture, training data, and design choices.
\section*{Limitations}

Our study has several important limitations. First, we observe that ModernBERT exhibits training instability during fine-tuning, this might be due to numerical instabitlity of the flashattention implementation.
Second, our downstream evaluation relies on established NLP benchmarks that may be reaching saturation, potentially masking more nuanced performance differences between architectures. Finally, our analysis focuses on base-sized models, and the relative performance characteristics may differ for larger model variants. Future work should address these limitations through stability analysis, development of more discriminative benchmarks, and evaluation across different model scales.

\section*{Ethics Considerations}

This work involves training large-scale language models using publicly available data, with special attention given to data quality, filtering, and documentation. We applied both heuristic and semantic filters to reduce harmful, biased, or low-quality content. Nonetheless, we acknowledge that pretrained models may still reflect societal biases present in the underlying data. We encourage responsible use of our models and welcome future research focused on auditing and mitigating bias and potential misuse.

\section*{Acknowledgments}

This work has received partial funding Benoît Sagot and Djamé Seddah’s chairs in the PRAIRIE-PSAI, funded by the French national agency ANR, as part of the “France 2030” strategy under the reference ANR-23-IACL-0008. This project also received funding from the BPI Scribe project.
The authors extend their gratitude to the OPAL infrastructure of Université Côte d'Azur for providing essential resources and support.
This work was also granted access to the HPC resources of IDRIS by GENCI under the allocation 2024-GC011015610 and AD011013900R2.  
Special thanks to Nathan Godey and Francis Kulumba for their assistance with training code and for engaging in productive discussions.

All pretraining dataset are sourced from publicly available and wildly used corpora (CultraX from OSCAR [CC0 1.0], and mc4 [odc-by], HAL's open archive [CC],  Wikipedia [cc-by-sa-3.0], RedPajamaV2 [Common Crawl Permissive license]). Downstream dataset are all from standard open benchmarks. ModernBERT's codebase is Apache-2.0. CamemBERTaV2, ModernCamembert checkpoints and models are released under the MIT licence.\footnote{\href{https://huggingface.co/collections/almanach/moderncamembert}{https://huggingface.co/collections/almanach/ moderncamembert}}

\bibliography{custom.bib}
\bibliographystyle{colm2025_conference}

\appendix
\onecolumn
\section{Quality Labeling Prompt}
\label{appendix:prompt}

\begin{tcolorbox}[title={Prompt used to annotate text quality}, label={prompt:llama-label}]
\begin{Verbatim}[breaklines=true, breaksymbolleft={}]
Below is an extract from a web page. Evaluate the quality of the content based on the following factors:

    1. Content Accuracy: Assess the correctness and reliability of the information presented. Consider the factual accuracy, use of credible sources (if mentioned), and absence of misinformation.
    2. Clarity: Evaluate how well the information is communicated. Look for clear explanations, well-defined terms, and logical flow of ideas.
    3. Coherence: Analyze the overall structure and organization of the content. Consider how well ideas are connected and if the content follows a logical progression.
    4. Grammar and Language: Assess the quality of writing, including correct grammar, spelling, and punctuation. Consider the appropriateness of language for the intended audience.
    5. Depth of Information: Evaluate the level of detail and thoroughness of the content. Consider whether it provides surface-level information or delves into more comprehensive explanations.
    6. Overall Usefulness: Assess the practical value and relevance of the information for a general audience. Consider how applicable or helpful the content would be for someone seeking information on the topic.

Based on these factors, give an overall quality score of low, medium, or high.

The extract:
    {input}

After examining the extract:
- Briefly justify your quality classification, up to 100 words on one line using the format: "Explanation: <justification>"
- Conclude with the quality classification using the format: "Quality score: <classification>" (on a separate line)

Remember to assess from the AI Assistant perspective, utilizing web search knowledge as necessary. Evaluate the content based on the quality factors outlined above.
\end{Verbatim}
\end{tcolorbox}

\section{Pretraining Details and Hyperparamters}
\label{appendix:pretraining-hp}

We closely follow the original ModernBERT recipe. We present the model parameters in Table~\ref{tab:model_hyperparams} and pretraining hyperparamters in Table ~\ref{tab:pretraining_hyperparameters}

\begin{table*}[!t]
\small
\centering
\begin{tabular}{lcc}
\toprule
Parameter & Base \\
\midrule
Vocabulary & 32,768 \\
Unused Tokens & 418 \\
Layers & 22 \\
Hidden Size & 768 \\
Transformer Block & Pre-Norm \\
Activation Function & GeLU \\
Linear Bias & False \\
Attention & Multi-head  \\
Attention Heads & 12  \\
Global Attention & Every three layers  \\
Local Attention Window & 128 \\
Intermediate Size & 1,152 \\
GLU Expansion & 2,304 \\
Normalization & LayerNorm \\
Norm Epsilon & 1e-5 \\
Norm Bias & False \\
RoPE theta & 160,000 \\
Local Attn RoPE theta & 10,000 \\
\bottomrule
\end{tabular}
\caption{ModernBERT model parameters}  
\label{tab:model_hyperparams}
\end{table*}

\begin{table*}[!t]
\small
\centering
\begin{tabularx}{\textwidth}{lccc} 
    \toprule
                    & Pretraining & Context Extension & Context Ext \& High Quality \\
    \midrule
    Training Tokens & 1 trillion & 150 billion & 100 billion \\
    Max Sequence Length & 1,024 & 8,192 & 8,192 \\
    \midrule
    Batch Size & 4,608 & 768 & 768 \\
    \hspace{3mm}Warmup (tokens) & \multicolumn{3}{l}{\hspace{1.2mm}None} \\
    \hspace{3mm}Microbatch Size & 96 & 8 & 8 \\
    \midrule
    Learning Rate & 8e-4 & 3e-4 & 3e-4 \\
    \hspace{3mm}Schedule & Trapezoidal & Trapezoidal & 1-sqrt (50B tokens delayed) \\
    \hspace{3mm}Warmup (tokens) & 3 billion & None & None \\
    Weight Decay & \multicolumn{3}{l}{\hspace{1.2mm}1e-5 } \\
    \midrule
    Training Time (hours) & 22.7 & 4.5 & 4 \\
    \midrule
    Model Initialization & Megatron & - & - \\
    \midrule
    Dropout (attn out) & \multicolumn{3}{l}{\hspace{1.2mm}0.1} \\
    Dropout (all other layers) & \multicolumn{3}{l}{\hspace{1.2mm}0.0} \\
    \midrule
    Optimizer & \multicolumn{3}{l}{\hspace{1.2mm}DecoupledAdamW} \\
    \hspace{1.2mm}Betas & \multicolumn{3}{l}{\hspace{1.2mm}(0.90, 0.98)} \\
    \hspace{1.2mm}Epsilon & \multicolumn{3}{l}{\hspace{1.2mm}1e-06} \\
    \midrule
    Training Hardware & \multicolumn{3}{l}{\hspace{1.2mm}48 GPUs - 12x(4xH100) } \\
    Training Strategy & \multicolumn{3}{l}{\hspace{1.2mm}FSDP - Full Sharding} \\
    Software Libraries & \multicolumn{3}{l}{\hspace{1.2mm}PyTorch 2.5.1, Cuda 12.4, Composer 0.28, Flash Attention 2.6.3-Hopper} \\
    \bottomrule
\end{tabularx}
\caption{Pre-training hyperparameters}  
\label{tab:pretraining_hyperparameters}
\end{table*}

\section{Detailed MTEB Scores}

\begin{table*}[!t]
\centering
\resizebox{\textwidth}{!}{
    \begin{tabu}{lcccccccc}
        \toprule
         & \textsc{Clustering} & \textsc{Classification} & \textsc{Pair Classification} & \textsc{Retrieval} & \textsc{Reranking} & \textsc{STS} & \textsc{Summarization} & \textsc{Overall} \\
        \midrule
        CamemBERTV2 & \textbf{39.40{\scriptsize$\pm$1.67}} & 62.40{\scriptsize$\pm$0.55} & 56.60{\scriptsize$\pm$0.55} & \textbf{39.40{\scriptsize$\pm$0.55}} & \textbf{65.20{\scriptsize$\pm$1.92}} & \textbf{75.60{\scriptsize$\pm$0.55}} & 31.00{\scriptsize$\pm$0.71} & \textbf{51.67{\scriptsize$\pm$0.57}} \\
        \midrule
        CamemBERTaV2 & 26.80{\scriptsize$\pm$1.92} & 41.00{\scriptsize$\pm$6.04} & 55.60{\scriptsize$\pm$3.21} & 6.80{\scriptsize$\pm$2.68} & 35.20{\scriptsize$\pm$3.83} & 52.00{\scriptsize$\pm$20.65} & 28.20{\scriptsize$\pm$1.92} & 31.15{\scriptsize$\pm$5.26} \\
        \midrule
        ModernBERT-CV2 & 39.20{\scriptsize$\pm$1.79} & \textbf{62.60{\scriptsize$\pm$0.55}} & \textbf{59.60{\scriptsize$\pm$1.14}} & 32.20{\scriptsize$\pm$3.70} & 56.80{\scriptsize$\pm$2.95} & 74.20{\scriptsize$\pm$1.30} & 29.80{\scriptsize$\pm$0.84} & 49.45{\scriptsize$\pm$1.36} \\
        ModernBERT-CV2-final & 39.20{\scriptsize$\pm$1.30} & 61.40{\scriptsize$\pm$0.55} & 59.40{\scriptsize$\pm$2.07} & 31.20{\scriptsize$\pm$0.84} & 55.20{\scriptsize$\pm$0.45} & 73.40{\scriptsize$\pm$0.55} & \textbf{31.20{\scriptsize$\pm$0.84}} & 48.79{\scriptsize$\pm$0.45} \\
        ModernBERT-HQ & 39.20{\scriptsize$\pm$0.84} & 62.00{\scriptsize$\pm$0.71} & 57.40{\scriptsize$\pm$0.89} & 34.20{\scriptsize$\pm$1.64} & 58.60{\scriptsize$\pm$1.52} & 75.00{\scriptsize$\pm$1.22} & 31.00{\scriptsize$\pm$0.71} & 49.93{\scriptsize$\pm$0.60} \\
        ModernBERT-HQ-final & 38.40{\scriptsize$\pm$0.55} & 60.00{\scriptsize$\pm$0.71} & 57.60{\scriptsize$\pm$0.89} & 33.80{\scriptsize$\pm$2.39} & 60.80{\scriptsize$\pm$2.39} & 74.80{\scriptsize$\pm$0.84} & 30.00{\scriptsize$\pm$1.22} & 49.29{\scriptsize$\pm$0.81} \\
        \bottomrule
    \end{tabu}
}
\caption{MTEB task type results. \textbf{Bold} indicates best score overall. Scores represent the average performance across all tasks within each task type category with standard deviations.}
\label{tab:mteb_task_type_results}
\end{table*}

\end{document}